\pdfoutput=1

\documentclass[11pt]{article}

\usepackage[final]{acl}
\usepackage{todonotes}
\usepackage{times}
\usepackage{latexsym}
\usepackage{multirow}
\usepackage{booktabs}
\usepackage{layouts}

\usepackage[T1]{fontenc}

\usepackage[utf8]{inputenc}

\usepackage{microtype}

\usepackage{inconsolata}

\usepackage{wrapfig,lipsum,booktabs}

\usepackage{graphicx}

\newcommand{\inlinemaketitle}{{\let\newpage\relax\maketitle}}
\usepackage{tcolorbox}
\usepackage{caption}
\usepackage{fancyhdr}

\title{KaPQA: Knowledge-Augmented Product Question-Answering}

\author{
 \textbf{Swetha Eppalapally\textsuperscript{1}},
 \textbf{Daksh Dangi\textsuperscript{1}},
 \textbf{Chaithra Bhat\textsuperscript{1}},
 \textbf{Ankita Gupta\textsuperscript{1}},
\\
 \textbf{Ruiyi Zhang\textsuperscript{2}},
 \textbf{Shubham Agarwal\textsuperscript{2}},
 \textbf{Karishma Bagga\textsuperscript{2}},
 \textbf{Seunghyun Yoon\textsuperscript{2}},
\\
 \textbf{Nedim Lipka\textsuperscript{2}},
 \textbf{Ryan A. Rossi\textsuperscript{2}},
 \textbf{Franck Dernoncourt\textsuperscript{2}}
\\
\\
 \textsuperscript{1}University of Massachusetts Amherst,
 \textsuperscript{2}Adobe Research
\\
}

\begin{document}
\pagestyle{fancy}
\fancyhead{} 
\fancyhead[LO,LE]{Accepted at the ACL 2024 Workshop on Knowledge Augmented Methods for NLP}

\inlinemaketitle
\begin{abstract}

Question-answering for domain-specific applications has recently attracted much interest due to the latest advancements in large language models (LLMs). However, accurately assessing the performance of these applications remains a challenge, mainly due to the lack of suitable benchmarks that effectively simulate real-world scenarios.
To address this challenge, we introduce two product question-answering (QA) datasets focused on Adobe Acrobat and Photoshop products to help evaluate the performance of existing models on domain-specific product QA tasks. Additionally, we propose a novel knowledge-driven RAG-QA framework to enhance the performance of the models in the product QA task. Our experiments demonstrated that inducing domain knowledge through query reformulation allowed for increased retrieval and generative performance when compared to standard RAG-QA methods. This improvement, however, is slight, and thus illustrates the challenge posed by the datasets introduced.

\end{abstract}

\section{Introduction}

The advancements in large language models (LLMs) led to exponential growth in domain-specific applications. Question Answering has emerged as one of the prominent domain-specific applications. As the demand for accurate and reliable QA systems increases, generic RAG-QA approaches often struggle to deliver satisfactory results within the specialized domains. This challenge has spurred active exploration in this area, with researchers employing various novel methodologies \cite{nguyen2024enhancing, setty2024improving, jiang2024enhancing, rackauckas2024rag} to improve QA systems. 

Additionally, training and evaluating these systems rigorously remains crucial. This trend underscores the critical need for domain-specific QA datasets to facilitate the training and evaluation of such systems. Notably, while efforts have been directed towards releasing datasets across prominent and expansive domains such as Medicine \cite{pal2022medmcqa, pampari2018emrqa}, Finance \cite{chen2021finqa,zhu2021tat}, and Legal \cite{zhong2019jecqa, chen2023equals}, there remains an apparent scarcity of such datasets in the area of software products.

To address this gap, our work investigates such industry-specific QA datasets, namely the Adobe HelpX datasets, and releases them for others to benchmark against and further improve their QA systems.
These datasets comprise of user queries and their corresponding answers pertaining to Adobe products, specifically Acrobat and Photoshop. By providing these benchmark datasets, we aim to offer valuable resources for assessing the performance of domain-specific RAG-QA systems. The datasets will be released after obtaining relevant permissions from Adobe.

Furthermore, we introduce a novel LLM-based Knowledge-Driven RAG-QA framework designed to seamlessly accommodate domain knowledge into RAG-QA systems. This framework leverages comprehensive knowledge bases for query expansion, thereby enhancing both retrieval and generation in domain-specific QA tasks.

Through extensive experimentation, we've determined that performing accurate retrieval over these datasets poses a unique challenge. Even introducing this concept of query augmentation using knowledge directly from the corpora only helped improve the model so much. This illustrates how these datasets are challenging ones - as even complex frameworks such as the one we've proposed, could only result in so much improvement. 

By contributing these datasets and proposing an innovative framework, we aim to advance LLM technology and its application in domain-specific QA tasks, ultimately improving user experiences and operational efficiency across various industries.

\section{Related Work}
\begin{figure*}[t]
  \centering
  \includegraphics[width=0.9\linewidth]{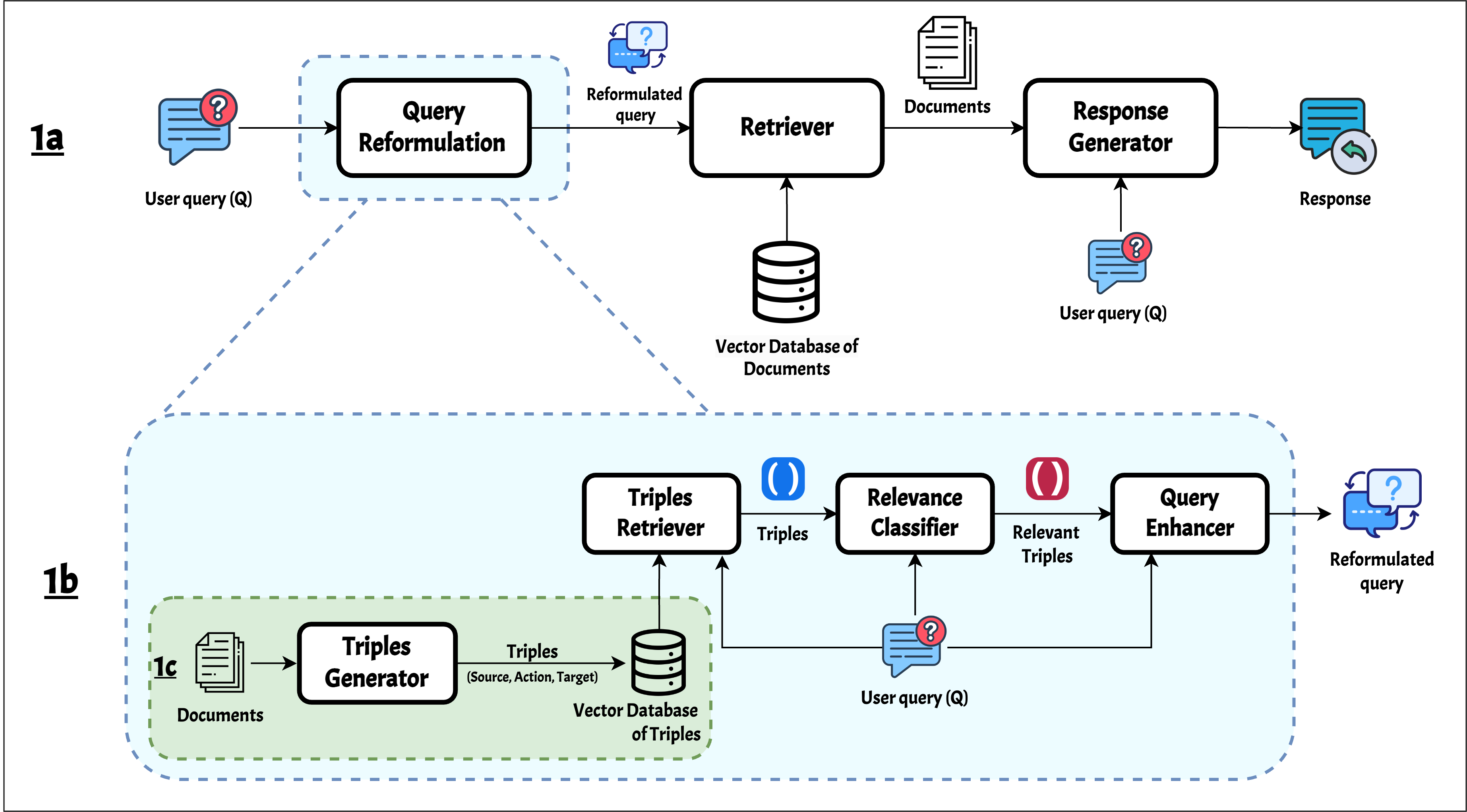} \hfill
  \caption {\small The figure represents our proposed framework. \textbf{\underline{1a}} depicts the main RAG-QA pipeline consisting of a retriever and a generator, along with our proposed query reformulation sub-pipeline. \textbf{\underline{1b}} gives a detailed view of the various components in our sub-pipeline. The process starts with the generation of knowledge base triples using the Triples Generator. Next, all matching triples to the user query are retrieved using the Triple Retriever, classified based on their relevance to the original query using the Relevance Classifier and finally reformulated using the Query Enhancer.}
  \label{fig:overview}
\end{figure*}
\subsection {Domain-specific question answering}
Several research efforts have been made to curate domain-specific question-answering benchmarks and training datasets, spanning domains like biomedical \cite{pal2022medmcqa, pampari2018emrqa, li-etal-2021-mlec}, finance \cite{chen2021finqa, zhu2021tat}, and legal \cite{zhong2019jecqa, chen2023equals}. In contrast, our work focuses on product question-answering, which is valuable in many enterprise settings. Furthermore, unlike many of these existing datasets that provide a simpler multiple-choice question-answer format, our work focuses on generative question-answering.

Among the research efforts in product-specific question-answering, \citet{yang2023empower} also provides a dataset focused on answering user queries about Microsoft products. However, many of the question-answer pairs in this dataset require a yes/no answer, with only a small portion requiring more complex answers. The PhotoshopQuiA~\cite{dulceanu2018photoshopquia} dataset is more closely related to our work in terms of domain, as it is also based on the Adobe Photoshop product. However, it specifically focuses on \textit{why} questions. In contrast, our work centers on \textit{how-to} queries, necessitating the model to generate a detailed sequence of steps to complete an operation. These answers are considerably more challenging to generate because their usefulness hinges on the accuracy of each individual step. If even one step in the generated answer is incorrect, the overall utility of the answer is compromised.

\subsection {Augmenting LLMs}
The Retrieval Augmented Generative (RAG) framework has been extensively worked on for years and \cite{gao2024retrievalaugmented, zhao2024retrievalaugmented, li2022survey} present a detailed examination of the progression of RAG paradigms \cite{ma2023query, Ilin_2024, shao2023enhancing, yu2023generate}, and introduce the metrics and benchmarks for assessing RAG models \cite{chen2023benchmarking, lyu2024crudrag}. One suggested future direction involves identifying methods to fully harness the potential of Large Language Models (LLMs) to enhance domain-specific RAG systems, aligning with our aim of leveraging LLMs to answer queries related to Adobe products.

Recent efforts aim to enhance LLMs' contextual generation in specific domains by incorporating external knowledge \cite{mialon2023augmented}. \citet{fatehkia2024trag} propose Tree-RAG where they utilize a tree structure to depict entity hierarchies in organizational documents and supplement context with textual descriptions for user queries. However, their approach is ineffective for documents lacking hierarchical organization such as ours. Another method to incorporate industry domain-specific information is presented by \citet{yang2023empower}. It involves getting a domain-specific language model with aligned knowledge and then feeding it to an LLM to generate enriched answers. We propose an alternative method to solve domain-specific RAG-QA through the construction of a comprehensive knowledge base consisting of triples and a multi-stage query reformulation pipeline. \citet{zhu2024llms} evaluates LLMs for Knowledge Graph (KG) tasks across diverse datasets, highlighting their suitability as inference assistants. Additionally, \citet{jagerman2023query} explore query expansion by prompting LLMs through zero-shot, few-shot and Chain-of-Thought (CoT) learning. Our work takes it a step further by incorporating knowledge base tuples in query expansion.
\section{Dataset Creation}

\subsection{Data Pre-processing}
The corpus is sourced from the publicly available Adobe HelpX\footnote{\url{https://helpx.adobe.com}} web pages for Acrobat and Photoshop products, which explain how to use the functionalities present in these products.

A crawling script is employed to extract the content from the web pages, segmenting them into distinct sections based on H2 headings. Each of these sections typically represents a specific topic or task within the respective product. The resulting sections tend  to be non-overlapping, facilitating targeted analysis. 

Throughout the process, all clickable and in-section links within the web pages are transformed into plain text to maintain consistency, while images are omitted to ensure that the corpus comprises solely textual content.

\subsection{Question-Answer Pairs with URLs Creation}
Gold question-answer (QA) pairs are meticulously crafted for analysis. Product experts, recruited through Upwork for Adobe Acrobat and Telus International, are instructed to write how-to questions and answers that provide procedural steps to accomplish specific tasks using the software. Furthermore, each QA pair is mapped with its respective source web page.

 The experts manually created question-answer pairs based on the respective HelpX web pages for Adobe Acrobat. Conversely, for Adobe Photoshop, GPT-4 initially generated question-answer pairs based on the web pages, which are subsequently reviewed and corrected by product experts to guarantee accuracy and relevance to the question and the answer. 
 
This systematic approach of the question-answer pair generation ensures the integrity and usability of the dataset for evaluation and research in the domain of software products and support.

\section {Data Analysis and Statistics} The Adobe Acrobat and the Photoshop datasets contain questions, answers, and corresponding source web page URLs. All questions in this dataset are how-to type asking steps to perform operations such as changing text font, editing text in a PDF, and creating certificate-based signatures. The gold answers to these questions provide procedural steps to accomplish these operations and the URLs of the web pages allow for independent verification of the answer. Table \ref{tab:statistics_dataset} provides insights into various metrics, such as the count of question-answer pairs, the average length of questions and answers, and the number of web pages and sections.  \\
\setcounter{footnote}{1}
\begin{table}[h]
\centering
\begin{minipage}{0.5\textwidth} %
\small
\begin{tabular}{@{}lll@{}}
\toprule
\textbf{Metric} & \textbf{Acrobat} & \textbf{Photoshop} \\ \midrule
No. of QA pairs & 131\footnote{\hsize=0.9\linewidth The count includes 22 composite questions. We focus on non-composite questions in this work (data analysis and experimentation). Composite Questions are an area for future exploration.}  & 96 \\ 
Avg. length of questions (words) & 8.80 & 12.74 \\
Avg. length of answers (words) & 118.71 & 98.71 \\
Total no. of web pages & 146  & 349 \\
Total no. of sections & 1281 & 2478 \\
Avg. no. of sections per web page & 8.78 & 7.1 \\
Avg. length of the section (words) & 135.75 & 121.09  \\
\bottomrule
\end{tabular}
\captionsetup{width=0.9\linewidth}
\caption{Statistics for the Adobe Acrobat and Photoshop datasets.}
\label{tab:statistics_dataset}
\end{minipage}
\end{table}
Answering these how-to queries presents significant challenges due to the critical nature of every step involved.  It's essential to emphasize that for an answer to the "How to" query to be entirely accurate, each step must be precise, and all steps must be in the correct sequence. Even a minor mistake in the explanation of a step or its order can invalidate the entire utility of the answer. Moreover, the average number of steps per response for a query in the Adobe Acrobat dataset is 4.71, indicating that most queries requires multi-step solutions. Below is an example highlighting the necessity for accurate and detailed instructions within every step of the response. \\
\begin{tcolorbox}[colback=white,colframe=black,title = Sample Q\&A from the dataset,label=box:exampleQA,height=7.2cm]
\textbf{Q:} How to insert Images into a PDF? \\
\textbf{A:} 
    1. Open the PDF in Acrobat and go to the \texttt{Edit} menu.
    
    2. Select \texttt{Image} from the \texttt{Add Content} submenu.
    
    3. In the dialog box, choose the image file you want to insert.
    
    4. Select the location where you want to insert the image or use the drag option to resize it as you insert.
    
    5. A copy of the image file will appear on the page with the same resolution as the original file.
\end{tcolorbox}
Adding to the complexity, a substantial portion of the Acrobat dataset, over 50\%, is comprised of implicit questions, defined by their brevity and conversational tone, lacking clear indications of user intent, as shown in Table \ref{tab:query_categories}. Furthermore, close to a quarter of the questions of the Acrobat dataset are ambiguous, lacking clear context and leading to multiple potential interpretations of the ask. Our proposed QA framework  has been designed considering these challenges and it effectively interprets the user ask. 

To assess the generalizability of our proposed QA framework, we curated an additional synthetic dataset focused around Adobe Photoshop product, closely resembling Adobe Acrobat in terms of question type, question and answer lengths, as shown in Table \ref{tab:statistics_dataset}. Additionally, the average number of steps per answer for Photoshop dataset is 4.6. However, since these are synthetically framed queries, they are well-formed, explicit, and unambiguous. By contrasting the characteristics of the synthetic dataset with those of Adobe Acrobat, we aim to evaluate the adaptability of our approach.

Moreover, both datasets serve as evaluation benchmarks, representing real-world user queries (both implicit and ambiguous) in Adobe Acrobat and controlled questions in the synthetic Adobe Photoshop dataset. They offer question-answer pairs for diverse scenarios, making them valuable resources for research in the software product domain.

\begin{table*}[h]
\centering
\begin{tabular}{|c|c|c|}
\toprule
\textbf{Category} & \textbf{Count of questions (\%)} & \textbf{Question Example} \\
 \midrule
\multirow{2}{*}{Explicit} & \multirow{2}{*}{48 (42.10\%)} & I need to increase image in PDF towards right direction, \\ 
& & how to do that in Acrobat? \\
Implicit & 66 (57.89\%) & resize jpg in Acrobat \\
Ambiguous & 28 (24.56\%) & Unable to delete PDF content need help. \\ \bottomrule
\end{tabular}
\caption{Question Categories with examples in the Acrobat dataset.}
\label{tab:query_categories}
\end{table*}

\section{Methodology}
As summarized by \cite{zhao2024retrievalaugmented, li2022survey, gao2024retrievalaugmented, lewis2021retrievalaugmented}, in a standard RAG-QA process, upon receiving an input query, the retriever identifies and retrieves pertinent data sources, which are subsequently utilized by the response generator to enrich the overall generation process. To enable Adobe domain-specific QA, we add an initial query reformulation stage which enhances the user query using knowledge base triples. Query reformulation or rewriting \cite{anand2023context, ma2023query} encompasses a set of techniques to transform a user's original query into one that's better aligned with the user's intent, thereby enhancing the retrieval outcomes. Our proposed query reformulation pipeline consists of multiple steps as shown in Fig.\ref{fig:overview}. It starts with the generation of knowledge base triples using the Triples Generator. Next, all matching triples to the user query are retrieved using the Triple Retriever, classified based on their relevance to the original query using the Relevance Classifier and finally reformulated using the Query Enhancer. Refer to Appendix~\ref{sec:appendix1} for the LLM prompts used at various stages. Given below is a detailed outline of the different components in our pipeline:
\\\\
\textbf{Step 1: Triples generation}. The goal is to represent each document as a collection of triples, each of which represents the key information contained within a document. Each triple is of the form (Source, Action, Target) to mimic what might be asked through a query i.e. assistance to act on a target. E.g., one of the generated triples from a document about editing text and text boxes is (rotation handle, rotate, text box) - the source of the action (rotate) is the rotation handle, which acts on a text box. Similarly, for each document, the LLM contextualizes input text using its vast knowledge base, identifying entities, actions, and relationships, before generating triples by selecting relevant phrases as sources, actions, and targets, informed by inferred context and linguistic patterns. The number of triples for each document varies (approximately 1 to 35 triples) based on the document's content and the model’s comprehension. Each triple is then encoded into a numerical vector representation using a pre-trained sentence encoder model which converts the textual elements of the triple into dense vectors. These are organized into a high-dimensional index structure, enabling efficient similarity search.
\\\\
\textbf{Step 2: Triples Retrieval}. This stage accepts the user query as input and then searches the vector store to retrieve all triples related to the query by calculating the similarity scores between the query vector and the vectors of stored triples, utilizing a similarity search algorithm. For every user query, it over-retrieves numerous triples. 
\\\\
\textbf{Step 3: Relevance Classification}. Through the previous step, we obtain numerous triples that have some relevance to the user query. In this stage, we use the capabilities of an LLM to identify only those triples that are the most relevant to the user query. The content of the document along with the list of the triples retrieved in Step 2 are passed to the LLM as a prompt with the instruction that it identifies and return only those triples that are the most relevant to the user’s query. Only the triples that are classified as relevant are considered in the subsequent steps.
\\\\
\textbf{Step 4: Query Enhancement}. Here the user query is reformulated to ensure that it has all the necessary information within it that can help the retriever fetch the correct associated documents. This reformulation is a form of query enhancement where the user query is augmented with words that are used interchangeably in the Adobe products domain. This gives more information for the retriever to use through which it can perform a more accurate search over its vector store and return the documents that are more likely to be relevant. The relevant triples along with the original user query are passed as the prompt to the LLM which rephrases the query.

\section{Experiments}

\begin{table*}[]
\centering \small
\begin{tabular}{lllllll}
\toprule
\multirow{2}{*}{} & \multicolumn{1}{c}{\multirow{2}{*}{Retriever}} & \multicolumn{1}{c}{\multirow{2}{*}{\begin{tabular}[c]{@{}c@{}}Query reformulation\\ method\end{tabular}}} & \multicolumn{1}{c}{\multirow{2}{*}{\begin{tabular}[c]{@{}c@{}}Retriever\\ Hit Rate \end{tabular}}} & \multicolumn{3}{c}{Answer Similarity Score} \\ \cmidrule{5-7} 
 & \multicolumn{1}{c}{} & \multicolumn{1}{c}{} & \multicolumn{1}{c}{} & \multicolumn{1}{c}{ROUGE-L} & \multicolumn{1}{c}{BERTScore} & \multicolumn{1}{c}{G-Eval} \\ \midrule \vspace{0.1cm}
\multirow{3}{*}{Baselines} & BM25 & None & 41.2\%& 0.301 & 0.835 &  0.412\\ \vspace{0.1cm}
 & DPR & None & 73.6\% & 0.409 & 0.859 & 0.574  \\ \midrule
Ours & DPR & \begin{tabular}[c]{@{}l@{}}\textbf{augmentation with} \\ \textbf{domain-specific triples}\end{tabular} & \textbf{74.7\%} & \textbf{0.416} & \textbf{0.860} & \textbf{0.578} \\ \midrule \vspace{0.1cm}
\multirow{2}{*}{Ablation}  &  DPR & \begin{tabular}[c]{@{}l@{}}via general purpose LLM\\ (w/o triple retriever and relevance classifier)\end{tabular} &  70.2\% & 0.393 & 0.855 & 0.562  \\  \vspace{0.1cm}
 & DPR & w/o triple relevance classifier &  65.7\% & 0.385 & 0.852 & 0.557 \\ \midrule
Oracle & DPR & \begin{tabular}[c]{@{}l@{}}w/o triple retriever \\ (triples obtained from gold documents)\end{tabular} & 80.7\% & 0.455 & 0.870 & 0.649 \\ 
 \bottomrule
\end{tabular}
\caption{Performance of baseline methods, our proposed method, and the ablation experiments over Acrobat test set. In all the above experiments, we use GPT3.5 as the LLM for triple generation and answer generation.  The semantic similarity scores are computed between the gold and the generated answer. Among ablations, in \textit{w/o triple relevance classifier} setting, we provide all retrieved triples to the query expansion model; in \textit{via general purpose LLM} setting, we only use LLM to reformulate query without including any triples. In the Oracle setting, \textit{w/o triple retriever}, we use gold documents to generate triples and provide them to the relevance classifier. This setting gives us an upper bound on the performance when correct knowledge triples are known. The reported hit rate is 
 the top-3 hit rate.
}
\label{table:results}
\end{table*}
We conducted a variety of experiments (as shown in Tables \ref{table:results} and \ref{table:results2}) on the Adobe datasets, where the main datasets formed the corpus of texts to be retrieved from, a selected few of which would be passed in as part of the prompt to the LLM; and the gold set was used for measuring performance. They ranged from using different retrievers to incorporating multiple components and techniques into the RAG-QA pipeline to achieve a performance gain. Throughout the processes we used GPT-3.5 0301 and GPT-4 0314 from Azure OpenAI. 

\subsection{Baselines}

\textbf{BM25 retriever + LLM:} We utilize a BM25 retriever to scan the document corpus and select the top k (k=3) most relevant documents based on the user's query. The relevance of each document is calculated by considering the frequency of query terms within the document and the document's length relative to the entire corpus. These selected documents then serve as contextual input for subsequent response generation tasks.\\\\
\textbf{DPR + LLM:} The Dense Passage Retrieval (DPR) method utilizes embeddings to represent passages and queries as vectors, which are then indexed using a similarity search algorithm. When a query is received, DPR compares the vector representation of the query with those of passages in the store, selecting the top k documents with the highest similarity scores as context for answer generation.\\\\
\textbf{General purpose LLM (or QR with no Triples):} We add an intermediate step of query reformulation using an LLM to the second baseline. The LLM is instructed to improve the original user query directly without any additional information, i.e. without the addition of domain knowledge into the query.

\subsection{Evaluation Methods}
We use different metrics to evaluate the performance of the two main components of the RAG-QA pipeline. For retrieval, we employ the Hit Rate or the number of times the Gold document was correctly retrieved as the metric. To measure the quality of the generated answers, multiple metrics are explored. As supported by \citet{yang2023empower}, ROUGE-L \cite{lin-2004-rouge} was utilized for measuring the lexical overlap between the generated answers and Gold answers, and BERTScore \cite{zhang2020bertscore} for calculating the semantic overlap by measuring the distances of embeddings between both the answers. Additionally, we incorporate G-Eval with GPT-4 \cite{liu2023geval} as an LLM-based metric to measure the similarity, leveraging its capability to provide a highly adaptable and versatile evaluation with human-like accuracy, enhancing the comprehensiveness of our evaluation approach. Apart from these, we also perform human evaluation to ensure correctness since there is a lack of good metrics for long-form answers \cite{yang2023empower, fan-etal-2019-eli5}.

\section{Results}

\subsection{Performance on the Adobe dataset}

\begin{table*}[]
\centering \small
\begin{tabular}{lllllll}
\toprule
\multirow{2}{*}{} & \multicolumn{1}{c}{\multirow{2}{*}{Retriever}} & \multicolumn{1}{c}{\multirow{2}{*}{\begin{tabular}[c]{@{}c@{}}Query reformulation\\ method\end{tabular}}} & \multicolumn{1}{c}{\multirow{2}{*}{\begin{tabular}[c]{@{}c@{}}Retriever \\ Hit Rate\end{tabular}}} & \multicolumn{3}{c}{Answer Similarity Score} \\ \cmidrule{5-7} 
 & \multicolumn{1}{c}{} & \multicolumn{1}{c}{} & \multicolumn{1}{c}{} & \multicolumn{1}{c}{ROUGE-L} & \multicolumn{1}{c}{BERTScore} & \multicolumn{1}{c}{G-Eval} \\ \midrule \vspace{0.1cm}
\multirow{3}{*}{Baselines} & BM25 & None & 79.17\% & 0.336 & 0.822 & 0.494\\ \vspace{0.1cm}
 & DPR & None & \textbf{92.70\%} & 0.470 & 0.879  & \textbf{0.828} \\ \midrule
Ours & DPR & \begin{tabular}[c]{@{}l@{}}\textbf{augmentation with} \\ \textbf{domain-specific triples}\end{tabular} & \textbf{92.70\%} & \textbf{0.480} & \textbf{0.880} & 0.776 \\ \midrule \vspace{0.1cm}
\multirow{2}{*}{Ablation}  &  DPR & \begin{tabular}[c]{@{}l@{}}via general purpose LLM\\ (w/o triple retriever and relevance classifier)\end{tabular} &  83.33\% & 0.406 & 0.859  & 0.692  \\  \vspace{0.1cm}
 & DPR & w/o triple relevance classifier &  91.67\% & 0.447 & 0.873 & 0.755 \\
 \bottomrule
\end{tabular}
\caption{Performance of baseline methods, our proposed method, and the ablation experiments over the Photoshop test set. In all the above experiments, we use GPT3.5 as the LLM for triple generation and answer generation. The semantic similarity scores are computed between the gold and the generated answer. The reported hit rate is the top-3 hit rate.}
\label{table:results2}
\end{table*}

Table \ref{table:results} presents the results for baseline methods, our proposed method, and the ablations of different components in our proposed method. We observe that our proposed method outperforms (Hit Rate: 74.7$\%$; GPTEval: 0.58) the baselines without any query reformulation when using BM25 (Hit Rate: 41.2$\%$; GPTEval: 0.41) and DPR retrievers (Hit Rate: 73.6$\%$; GPTEval: 0.57). Among the baseline retrievers, the DPR-based method performs better than the BM25 retrieval. Therefore, we present other results using the DPR retriever.

Our method also performs better than the baseline using simple LLM prompting (i.e., without including any domain-specific knowledge) to reformulate the query (Hit Rate: 70.2$\%$; GPTEval: 0.56). On qualitative examination of reformulated queries, we observe that while query reformulation using a general-purpose LLM can make queries more grammatical or well-formed, it still cannot augment the queries with domain-specific knowledge. On the other hand, our method of using LLM-generated triples can help link entities that have similar meanings within the domain. 
For instance, for the query "How to convert word docs to pdf," our method retrieves triples that aid in reformulating the query to "How to convert files to pdf," thereby assisting in retrieving the correct document.

\subsection{Ablation Studies}
Next, we perform a series of ablations on our framework to evaluate the functions of various components in our pipeline.

\textbf{Ablation on relevance classifier:} 
We directly provide all the triples retrieved by the triples retriever to the query expansion model without filtering out any retrieved triples via the relevance classifier. This approach performs worse than our proposed model (Table \ref{table:results}), since the retrieved triples may include numerous noisy ones, which are then integrated into the query through the query expansion model, resulting in a noisy query. Thus, our results highlight the necessity for the relevance classifier to provide only highly relevant triples to the query expansion model.

\textbf{Ablation on Triple Retriever:} 
In this experiment, we only consider gold documents annotated for a given query in the Adobe dataset and generate triples from them. The generated triples are filtered via the relevance classifier and then fed to the query enhancer. As shown in Table \ref{table:results}, this setting performs the best due to the use of highly relevant triples for query reformulation. Our results highlight the importance of building an efficient triple retriever to improve overall performance.

\subsection{Performance on the Photoshop dataset}
Table \ref{table:results2} presents the results over the Photoshop test set. Once again, we observe that our proposed method is on par with the baseline, with a slight decrease in the GEval score. We attribute this to the fact that the Photoshop test set queries are already properly formulated; and reformulating these queries results in a slight deviation, hence the decrease in the G-Eval score. This illustrates that the performance of our proposed method is not dataset-specific. The pipeline is flexible enough to incorporate multiple different corpora - having created the respective knowledge bases.

\begin{figure}
    \includegraphics[width=0.48\textwidth]{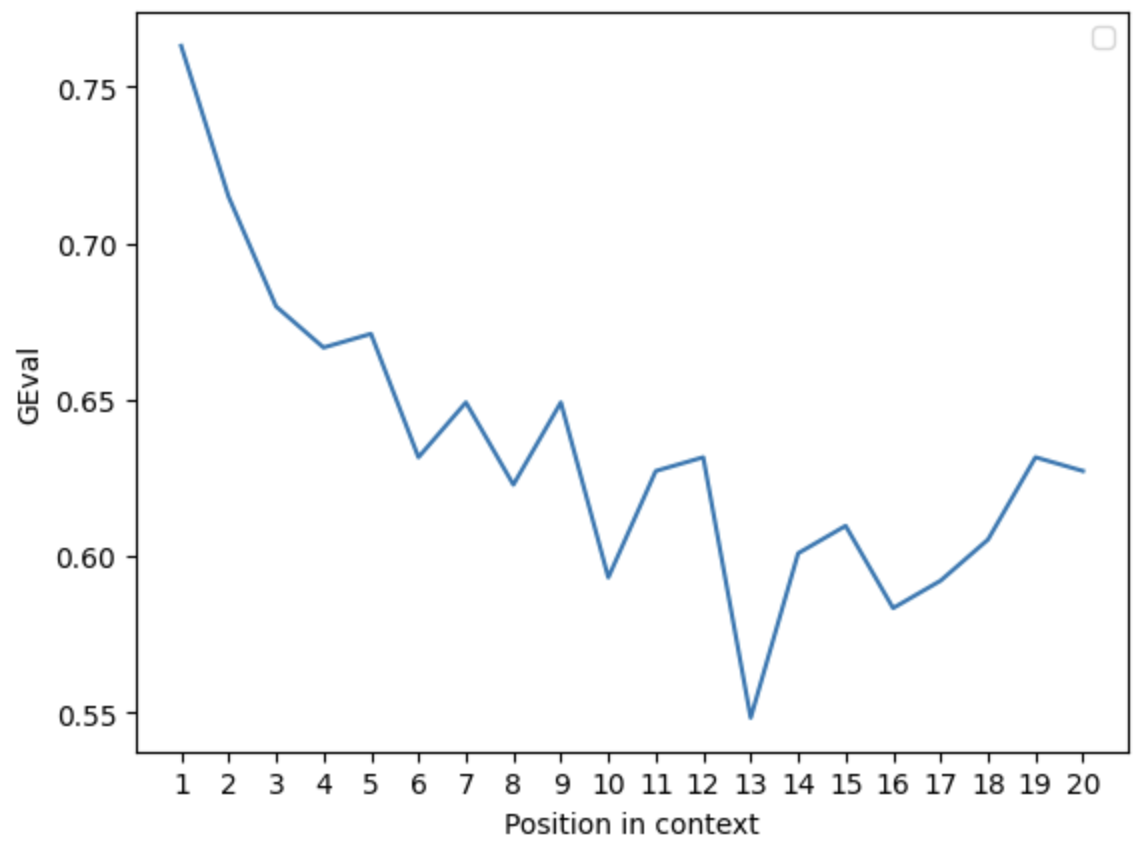}
    \captionof{figure}{GEval score relative to the position the gold document is passed in as context over Acrobat test set.}
    \label{fig:ndcg_exp}
\end{figure}

\subsection{Error analysis}
\begin{figure}
    \includegraphics[width=0.5\textwidth]{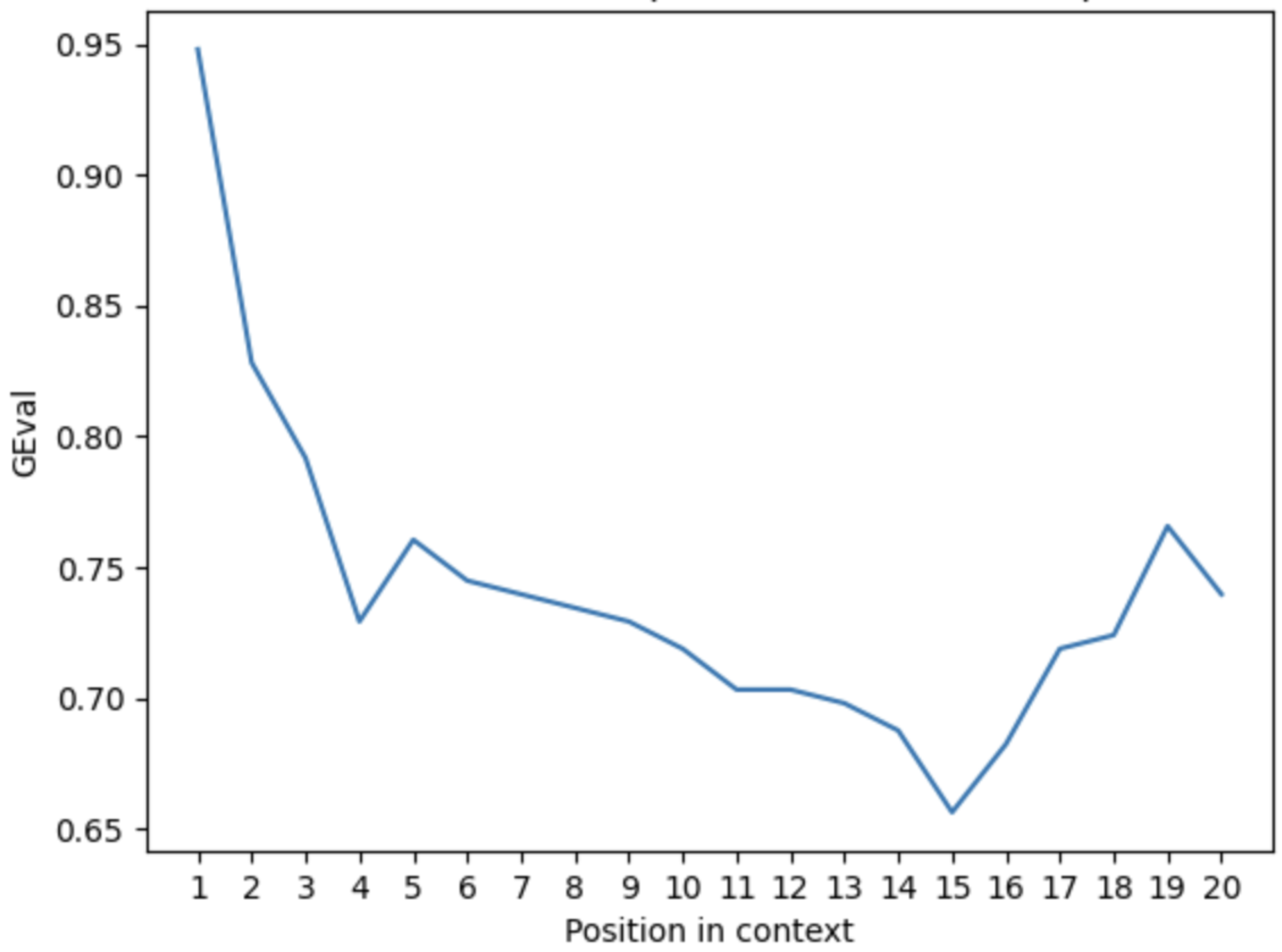}
    \captionof{figure}{GEval score relative to the position the gold document is passed in as context over Photoshop test set.}
    \label{fig:ps_ndcg_exp}
\end{figure}

While our method is able to link entities that are related in a particular domain, we also observe some cases of errors. For instance, the query `Create PDFs of specific size by cutting a large PDF into a smaller file size.' was reformulated to `How can I reduce the size of a PDF file?' as a result of retrieving the following triples : \\
1. (Reduce File Size command, reduces, size of PDF), 2. (PDF, reduce, size), 3. (PDF Optimizer, reduces, size of PDF files). These triples seemed to focus more on the keyword "size" and seemed to attribute the word "cutting" to reducing, while the intention of the original query was more towards splitting a PDF into multiple PDFs. This reformulation seemed to mis-interpret the original question, which resulted in incorrect retrieval. We attribute the source of such error to the noise originating from the retrieved triples that are irrelevant to the user query. Our analysis further highlights the importance of a high-precision triple retriever, as also suggested by our ablation study, which excludes the triple retriever and relies solely on triples from gold documents. Furthermore, another cause for concern was the  similarity score metrics only slightly increasing when compared to their retrieval metric counterparts. We realized that hit rate may not offer a holistic understanding of our retrievers performance. \cite{ravaut2024context} suggests that LLM's are sensitive and exhibit different utilization of input tokens depending on their position within the context provided, which we also found to be the case as shown in Figure \ref{fig:ndcg_exp} \& \ref{fig:ps_ndcg_exp}. It illustrates that the rank, or position, of the retrieved gold document impacts generation to a large extent, and is thus more indicative of proper retrieval than just hit rate alone. Following this, we decided to also consider NDCG as an evaluation metric, and the results are shown in Table \ref{tab:NDCG_vals}.

\begin{table}[h]
\centering
\begin{minipage}{0.5\textwidth}
\small
\begin{tabular}{@{}ll@{}}
\toprule
\textbf{Model} & \textbf{NDCG}  \\ \midrule
Query Reformulation w/ Triples(Proposed Model) & 0.447 \\
Query Reformulation w/o Triples & 0.453 \\
No Query Reformulation (DPR Baseline) & 0.507
\end{tabular}
\caption{NDCG values for different models.}
\label{tab:NDCG_vals}
\end{minipage}
\end{table}

\subsection{Performance using GPT-4o}
\begin{figure}[!htb]
    \includegraphics[width=0.45\textwidth]{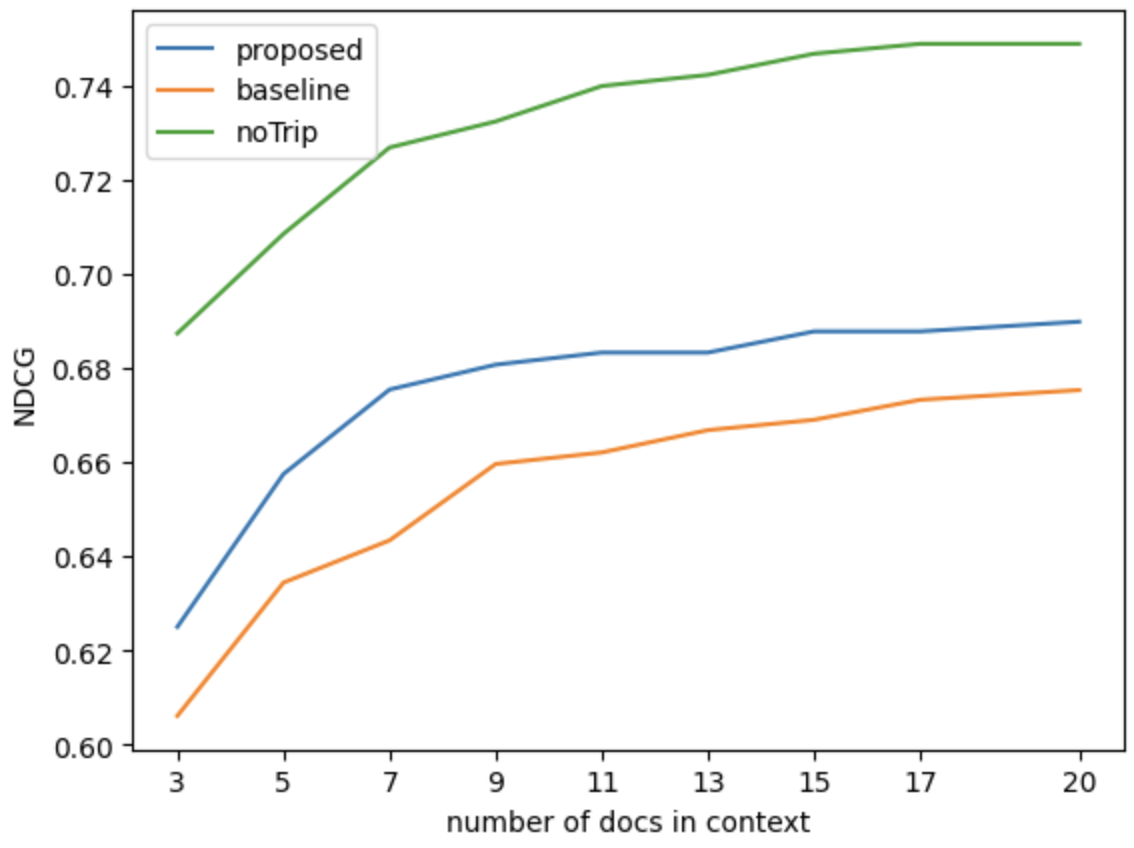}
    \captionof{figure}{NDCG scores for our proposed model, the DPR baseline, and query reformulation without triples using an LLM (noTrip) over the Acrobat test set.}
    \label{fig:gpt4o_ndcg}
\end{figure}

\begin{figure}[!htb]
    \includegraphics[width=0.45\textwidth]{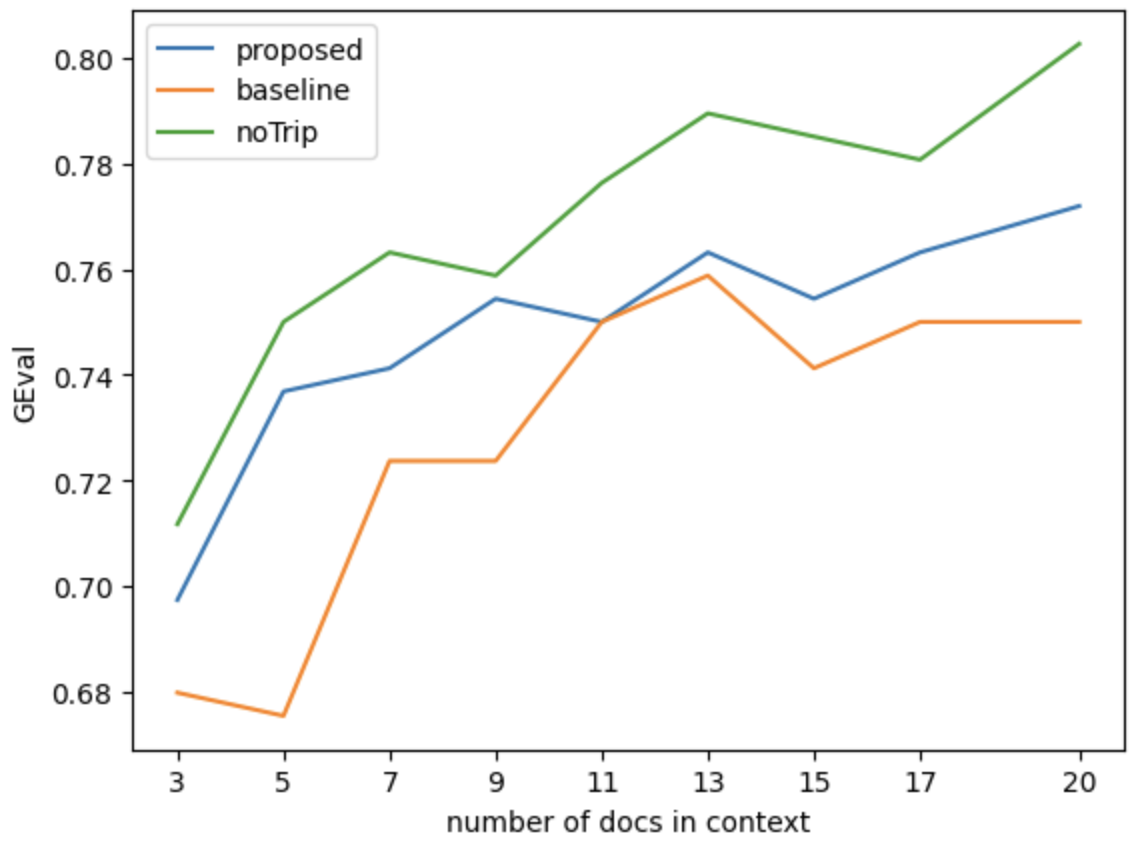}
    \captionof{figure}{GEval scores for our proposed model, the DPR baseline, and query reformulation without triples using an LLM (noTrip) over the Acrobat test set.}
    \label{fig:gpt4o_geval}
\end{figure}

Finally, we test our framework using a state-of-the-art model, GPT-4o, as well as different embeddings for the retriever to see how the performance would translate. In this experiment, we pass in a varying number of retrieved documents, and observe the corresponding NDCG and GEval values. Figures \ref{fig:gpt4o_ndcg} and \ref{fig:gpt4o_geval} present these results over the Acrobat test set. While the proposed model still outperforms the baseline in both of these metrics, vanilla query reformulation without triples seems to greatly outperform the aforementioned. Upon further qualitative analysis it was inferred that while GPT3.5 was more lax on introducing information from the triples into the query; GPT-4o tried to incorporate all of the information, which resulted in a far nosier query, thus leading to poorer retrieval.

Conversely, we observe the opposite results over the Photoshop dataset, as shown in Figures \ref{fig:ps_gpt4o_ndcg} and \ref{fig:ps_gpt4o_geval}, which indicate that the baseline outperforms the proposed models in this setting. We attribute this to the fact that the Photoshop test set was synthesized using LLMs, which resulted in accurately formulated queries. In such scenarios, the proposed model under performs the baseline, as it tries to reformulate the query and deviates too much from the original meaning in doing so, thus resulting in poorer retrieval.

\section{Conclusion}
In this paper, we introduce two QA datasets focused on Adobe Acrobat and Photoshop products, that serve as benchmarks to evaluate an RAG-QA framework tailored for domain-specific procedural long-form QA datasets. We also introduce a novel and detailed pipeline with components directed to improve the retrieval and generation metrics in such QA datasets. It equips LLMs with domain-specific knowledge through the use of Knowledge Base triples to bridge the gap between general RAG-QA methods and industry demands. Through various experiments we showcase the effectiveness and limitations of our proposed pipeline in standard metrics.

\section{Limitations}
This research presents multiple opportunities for improvement that can be explored further. To enhance the versatility and applicability of our method, it would be advantageous to explore its effectiveness across a broader range of larger industry-specific datasets beyond those provided by Adobe. Additionally, although our RAG-QA framework exhibits advancements over the baselines and offers valuable insights, the noise introduced by the LLM during query reformulation makes room for enhancement in the retrieval process. Moreover, expanding the application of our framework beyond text-based QA scenarios to include multi-modal capabilities opens up exciting new possibilities and broadens its potential impact. Lastly, it's important to note that assessing long-form QA remains an ongoing area of research, highlighting the necessity for a well-defined and automated metric to ensure accurate evaluation. Addressing these limitations is crucial for advancing the efficiency and scope of future research endeavors.

\section{Acknowledgments}
The authors of this paper would like to thank Hamed Zamani and more generally the entire UMass COMPSCI 696DS course staff for their insightful feedback, as well as the anonymous reviewers for their constructive comments.

\bibliography{custom}
\appendix
\section{Code Release}
We will release the code along with the data at \url{https://github.com/daksh-dangi/KaPQA} upon getting the required permission from Adobe.

\newpage

\section{Prompt Structures}
\label{sec:appendix1}
In this section we have listed the high-level structure of all the prompts used for the components of our pipeline.

\begin{tcolorbox}[width=3.2in,
                  boxsep=2pt,
                  left=2pt,
                  right=2pt,
                  top=2pt,
                  label=box2,
                  title=Triples Generator LLM prompt
                  ]%
  System:\\
  You are an assistant for the Adobe Acrobat application that helps create tuples of the form (source, action, target) based on the information given to you.
  \\\\
  User:\\
  You are given a section from an adobe help document. Extrapolate the most relevant relationships you can from the context and generate tuples of the form (source, action, target). Ensure that the sources, actions, and targets are directly present in the provided context.\textbackslash n\\
  You must use only the provided data in variable 'Context' to identify relationships.\textbackslash n\\
  You must not use any other information from any other source or from previous knowledge beyond the provided 'Context'.\textbackslash n\\
  Example: If the document contained the phrase "To edit the image, first click on the triple line menu", one relevant tuple would be (triple line menu, edit, image). Here, the source of the action (editing the image) is the triple line menu. The direct effect of this action is on the image, hence that is the target. In a similar manner, create tuples for the provided context.\\
  Context: <CONTEXT>\\
  Constraints:\textbackslash n\\
  1. The created tuples must form the same format as the example provided.\textbackslash n\\
  2. The source and target in the tuple must only reference objects or menu items, and no actions\textbackslash n\\
  3. Ensure that the tuples generated are all related to the title of the section: <SECTION HEADER>\textbackslash n\\
  4. Only generate the most relevant tuples for the provided document with the given section header\textbackslash n\\
  5. You must make the contents of the tuples short and concise\textbackslash n\\
  6. Ensure the words "Adobe", "PDF", and "Acrobat" are not in the generated tuples.\textbackslash n
\end{tcolorbox}

\begin{tcolorbox}[width=3.0in,
                  boxsep=2pt,
                  left=2pt,
                  right=2pt,
                  top=2pt,
                  label=box3,
                  title=Relevance Classifier LLM prompt
                  ]%
  System:\\
  You are an assistant for the Adobe Acrobat application. You are designed to filter out the information provided and classify what is most relevant to the given query.
  \\\\
  User:\\
  A user has provided you with the following query: <USER QUERY>\textbackslash n\\
  Use the data given in the variable 'Context' to classify which of the data elements are most relevant to the user's query.\textbackslash n\\
  Data in the 'Context' variable is of the form (source, action, target), where the first value contains the source of the action that is directed towards the target.\\
  Example: One data element could be (triple line menu, edit, image). If the query was asking about how to edit an image, this element would be relevant. However, if the query was instead asking how to edit a video, it would be very irrelevant, and hence should not be included. The presence of a query word in the tuple does not make it relevant, look at the meaning as well when you are considering relevance. In a similar manner, filter out the context for the provided provided document.\\
  Context: <CONTEXT>\\
  Constraints:\textbackslash n\\
  1. Retrieve the data elements that are most relevant to the action that the user is trying to do in the query provided.\\
  2. Ensure that the source and target of the data elements retrieved are similar to what is present in the query. \textbackslash n\\
  4. Give the most relevant data elements in a numbered list, and only provide the data elements themselves. No explanation.\textbackslash n
\end{tcolorbox}

\begin{tcolorbox}[width=3.2in,
                  boxsep=2pt,
                  left=2pt,
                  right=2pt,
                  top=2pt,
                  label=box4,
                  title=Query Enhancer LLM prompt
                  ]%
  System:\\
  You are an assistant that is designed to only enhance user queries.
  \\\\
  User:\\
  You are given a query by the user and you must enhance the query by only using the data provided in variable 'Tuples'. The 'Tuples' variable is of the form (source, action, target).\textbackslash n\\
  Constraints:\textbackslash n\\
  1. Rephrase the query using the provided tuples, but do not change the meaning of the initial query.\textbackslash n\\
  2. Only use information from the tuples that are relevant to the query to reform the query.\textbackslash n\\
  3. Make the rewritten query one sentence at most.\textbackslash n\\
  4. Re-write the query in a manner similar to how a human might search for an answer on a help page. Keep the query short.\textbackslash n\\
  5. Only reformulate the given query, without answering it.\textbackslash n\\
  6. You must not use any other information from any other source or from previous knowledge beyond the provided 'Tuples'.\textbackslash n\\
  7. Ensure the words "Adobe" and "Acrobat" are not in the query.\textbackslash n\\
  8. Only answer with one reformulated query. Example:\textbackslash n\\
  Given Query: 'how to remove letters from a text box'\textbackslash n \\
  Tuples: (text, delete key, remove),(page thumbnail, delete key, remove), (text, font item, edit), (text, font item, remove)\textbackslash n\\
  Reformulated Query: how to delete text\\
  In a similar manner, reformulate the query below.\\
  Given Query: <USER QUERY>\\
  Tuples: <Tuples>\\
  Reformulated Query:
\end{tcolorbox}

\begin{tcolorbox}[width=3.0in,
                  boxsep=2pt,
                  left=2pt,
                  right=2pt,
                  top=2pt,
                  label=box1,
                  title=DPR + General purpose LLM Prompt
                  ]%
  System:\\
  You are an assistant for the Adobe application that is designed to only enhance user queries. When asked about anything that does not relate to Adobe, only reply with 'Content not found'
  \\\\
  User:\\
You are asked a question by the user and you must enhance the query. Do not answer the query, only change it's wording.\textbackslash n\\
You must not use any other information from any other source or from previous knowledge beyond the query provided.\textbackslash n\\
Understand what might be the cause of confusion, and rewrite the query by trying to model what the user could have been asking.\textbackslash n\\
Ensure that the reformulated query is bound by the given constraints.\textbackslash n\\
Query to be enhanced: "{query}"
Constraints:\textbackslash n\\
1. Make the rewritten query one sentence at most.\textbackslash n\\
2. Make sure that the rewritten query does not have any excessive adjectives, and is short and to the point.\textbackslash n\\
3. Only reformulate it, without answer it.\textbackslash n\\
4. Only answer with the reformulated query.
\end{tcolorbox}

\newpage
\section{Graphs of Evaluation Metrics across models using GPT-4o}

\begin{figure}[!htb]
    \includegraphics[width=0.45\textwidth]{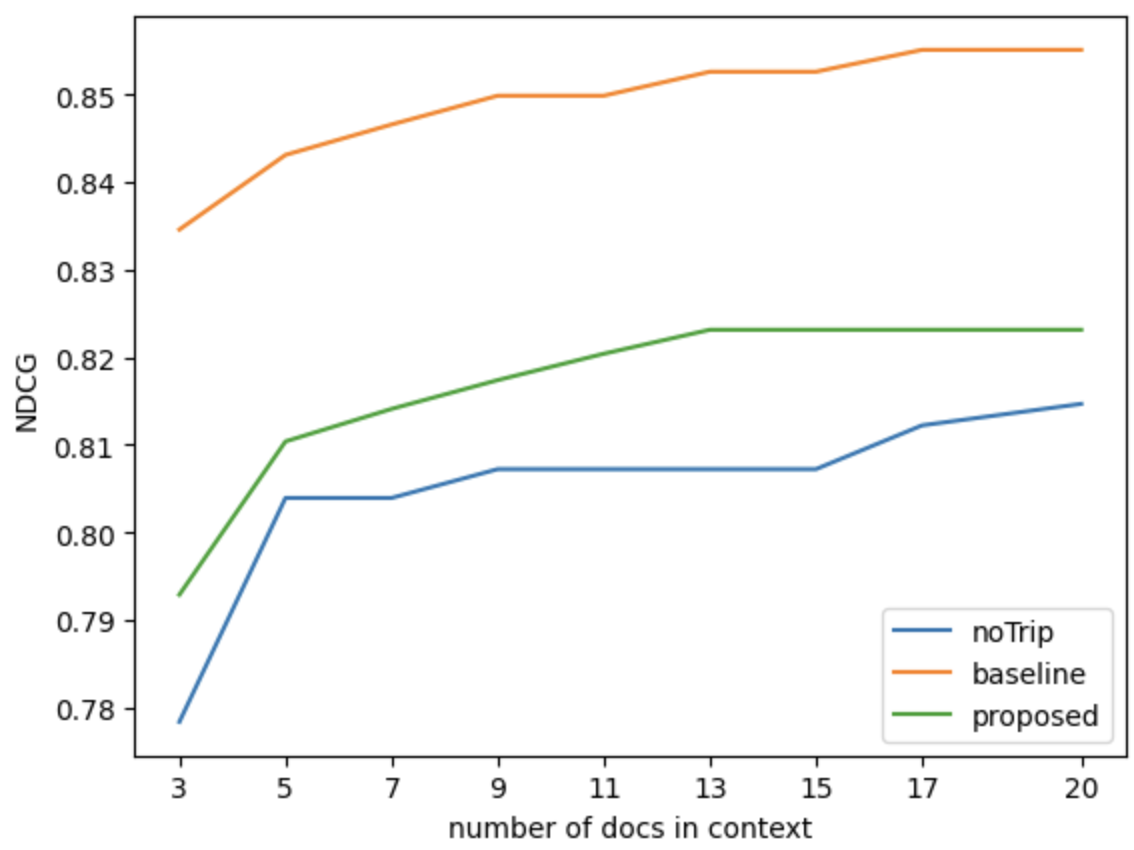}
    \captionof{figure}{NDCG scores for our proposed model, the DPR baseline, and query reformulation without triples using an LLM (noTrip) over the Photoshop test set.}
    \label{fig:ps_gpt4o_ndcg}
\end{figure}

\begin{figure}[!htb]
    \includegraphics[width=0.45\textwidth]{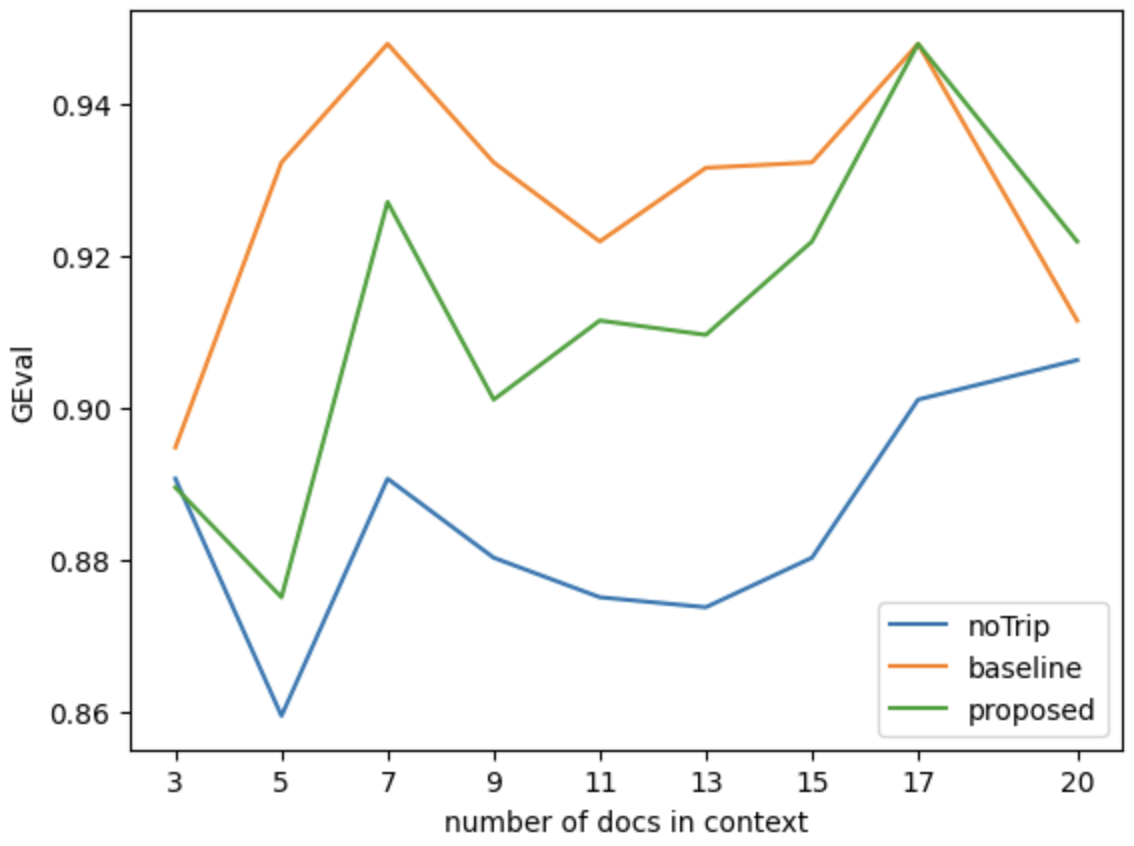}
    \captionof{figure}{GEval scores for our proposed model, the DPR baseline, and query reformulation without triples using an LLM (noTrip) over the Photoshop test set.}
    \label{fig:ps_gpt4o_geval}
\end{figure}

\section{Examples of proposed method outperforming DPR + GPT3.5.}
\label{sec:appendix1}
We have listed some examples in Table \ref{table:examples1} where the proposed model using query reformulation outperforms the DPR + GPT3.5 baseline. \\

\begin{table*}[]
\centering \small
\begin{tabular}{p{0.14\linewidth} | p{0.14\linewidth} | p{0.20\linewidth} | p{0.20\linewidth} | p{0.19\linewidth}}
\toprule
Original query & Reformulated Query & Generated Answer without QR & Generated Answer with QR & Gold Answer \\
\midrule
test box resizing in acrobat & How to resize a text
box or text in a PDF using the Edit panel in
Acrobat? & To resize a box in Acrobat, follow these steps: 1. Open the PDF file in Acrobat.2. Choose the hamburger Menu (Windows), or the File menu (macOS)
> Print. 3. Select Size, and choose one of the
following options: - Scales small pages up and
large pages down to fit the paper. %
[...] 4. If you don’t see a Page Scaling options,
click the arrow next to the Printer menu to expose
more controls in the Print dialog box.& To resize a text box in Acrobat, follow these steps: 1. Select the text box you want to resize by clicking on it. 2. Drag one of the selection handles on the edge of the box to resize it. 3. To maintain alignment with other list items, press Shift as you drag. That's it! You have successfully resized the text box in Acrobat. & 1. Choose `Edit` from the mega verb bar. The PDF switched is to edit mode and the Edit PDF panel will be displayed. 2. Click on an edge of the text box to select it. 4. Drag any of the handles that appear on the corners or edges of the text box to resize it. \\
\midrule
adding image on a pdf file & What are the different ways to add images to a PDF file? & To add an image to a PDF file, follow these steps: 1. Open the PDF file in Acrobat. 2. Choose Edit from the menu and select the image or object you want to edit. 3. If you want to edit multiple items together in the same file, select them all. 4. Under Adjust Objects in the Edit panel, select Edit using, and then choose an editor. 5. Make the desired changes in the editor. 6. Choose File > Save in the editor to save the changes. 7. The object is automatically updated and displayed in the PDF when you bring Acrobat to the foreground. If you want to copy and paste an image [...] file or a clipboard. & To add an image to a PDF file, follow these steps: 1. Open the PDF file in Acrobat. 2. Select Edit PDF\" from the right-hand pane. 3. Click on Image in the toolbar. 4. Select the image file you want to insert. 5. Choose where you want to insert the image or drag to resize it as you insert. 6. Once the image is added, you can move or resize it by hovering over the image and using the move or resize icons. 7. To adjust the image further, select Image from the left panel, [...] select File > Save. & 1.Open the PDF in Acrobat and go to the `Edit` mega verb. 2. Select `Image` from the `Add Content`. 3. In the dialog box that appears, choose the image file you want to insert. 4. Select the location where you want to insert the image or use the drag option to resize it as you insert. 5. A copy of the image file will appear on the page with the same resolution as the original file. \\
\midrule
Are we allowed to crop images in a pdf file? & Can images be edited and added to a PDF file? & Yes, you can crop images in a PDF file using Adobe Acrobat. Here are the steps to do so: 1. From the top tools bar, select Edit. Alternatively, select the All tools menu > Edit a PDF.\textbackslash n2. It displays the Edit toolset in the left panel. 3. From the Edit tools menu, select Crop Pages. 4. Drag a rectangle on the section that you want to crop. If necessary, drag the corner handles of the cropping rectangle until the page is the size you want.\textbackslash n5. Double-click [...] \textbackslash n8. Select OK. & Yes, you can crop images in a PDF file using Adobe Acrobat. Here are the steps to do so: 1. Open the PDF in Acrobat and select Edit. 2. Choose the image you want to crop. 3. Under Adjust Objects in the Edit panel, select Crop Image. 4. Drag the cropping handles to adjust the size of the image. 5. Click outside the image to apply the crop. 6. Save the PDF. & 1.Open the PDF in Acrobat and go to the Edit mega verb.\textbackslash n2. select the image you want to crop.\textbackslash n3. In the Adjust Objects section of the Edit panel, choose the Crop image tool.\textbackslash n4. A selection handle will appear around the image. Drag the selection handles to adjust the cropping area.\textbackslash n5. Once you have selected the desired area to crop, click outside the selection to apply the crop.\textbackslash n The image will be cropped to the selected area.\\ \bottomrule
\end{tabular}
\caption{Examples of proposed method outperforming DPR + GPT3.5.}
\label{table:examples1}
\end{table*}

\section{Examples of proposed method under performing DPR + GPT3.5}
\label{sec:appendix1}
We have listed some examples in Table \ref{table:examples2} where the proposed model using query reformulation under performs the DPR + GPT3.5 baseline.
\begin{table*}[]
\centering \small
\begin{tabular}{p{0.14\linewidth} | p{0.14\linewidth} | p{0.20\linewidth} | p{0.20\linewidth} | p{0.19\linewidth}}
\toprule
Original query & Reformulated Query & Generated Answer without QR & Generated Answer with QR & Gold Answer \\
\midrule
How can I create PDFs from all non-pdfs stored in a directory? Is it possible to do it in one go. & Is it possible to convert all non-PDF files in a folder to PDFs with electronic features using Adobe Acrobat? & To create PDFs from all non-PDFs stored in a directory in one go, follow these steps:\textbackslash n\textbackslash n1. In Acrobat, select Create from the global bar.\textbackslash n2. The Create a PDF tool opens. On the left rail, select Multiple files > Create multiple PDF files, and then select Next.\textbackslash n3. Choose Add Files > Add Files or Add Folders, and then select the files or folder.\textbackslash n4. Select OK. The Output options dialog box appears.\textbackslash n5. In the Output Options dialog box, specify your target folder and filename preferences, and then select OK.[...] & To create PDFs from all non-PDFs stored in a directory, you can follow these steps:\textbackslash n\textbackslash n1. In Acrobat, select Create from the global toolbar.\textbackslash n2. The Create a PDF tool opens. On the left rail, Single file is selected by default. Choose the Select a file button.\textbackslash n3. In the Open dialog box, navigate to the directory where your non-PDF files are stored.\textbackslash n4. Select all the non-PDF files you want to convert to PDF.\textbackslash n5. Select Create to convert the files to PDF.\textbackslash n6. Depending on the type [...] using this method. & 1. You can create multiple PDFs from multiple native files, including files of different supported formats, in one operation. %
[...] \textbackslash n1. In Acrobat, select Create from the global bar.\textbackslash n2. The Create a PDF tool opens. On the left rail, select Multiple files > Create multiple PDF files, and then select Next.\textbackslash n3. Choose Add Files > Add Files or Add Folders, and then select the files or folder.\textbackslash n4. Select OK. The Output options dialog box appears.\textbackslash n5. Specify your target folder and filename preferences, and then select OK. \\
\midrule
I need to create a link on a PDF. & How can I add a link to a PDF document? & To create a link on a PDF, follow these steps:\textbackslash n\textbackslash n1. In the source document, choose Edit > Link > Add or edit a link.\textbackslash n2. Drag a rectangle to specify a location for the link.\textbackslash n3. In the Create Link dialog box, set the link appearance.\textbackslash n4. Select Go to a page view, and then select Next.\textbackslash n5. In the target document, [...] create a link. & To create a link on a PDF, follow these steps:\textbackslash n\textbackslash n1. Choose the Select tool.\textbackslash n2. Position the pointer over the linked area on the page until the pointer changes to the hand with a pointing finger. A plus sign (+) or a w appears within the hand if the link points to the web. Then click the link.\textbackslash n3. In the Create Link [...] for a link to work correctly. & 1. Choose Edit > Link > Add or edit a link. The pointer becomes a cross-hair, and any existing links in the document, including invisible links, are temporarily visible.\textbackslash n2. Drag a rectangle where you want to create a link. This is the area in which the link is active.\textbackslash n3. In the Create Link dialog, choose the options you want for the link appearance.\textbackslash n4. Select one of the following link actions:\textbackslash n - Go to a page view\textbackslash n Select Next [...] associated with the link.\\
\midrule
How to remove password from the PDF & How to remove password protection from a PDF? & To remove password from a PDF, follow these steps:\textbackslash n\textbackslash n1. Open the PDF in Acrobat.\textbackslash n2. Go to All tools > Protect a PDF > Remove security.\textbackslash n3. If prompted, type the permissions password. If you don't know the permissions password, contact the author of the PDF.\textbackslash n4. Acrobat will remove the security from the PDF and you will be able to access it without a password. & To remove password from a PDF, follow these steps:\textbackslash n\textbackslash n1. Go to All tools > Protect a PDF > Protect with password from the Protect a PDF toolset in the left pane.\textbackslash n2. Choose Editing.\textbackslash n3. Type and retype your password. Your password must be at least six characters long. The password strength is displayed next to your password [...] using the password. & You can remove security from an open PDF if you have the permissions to do so. If the PDF is secured with a server-based security policy, only the policy author or a server administrator can change it.\textbackslash n\textbackslash n1. Open the PDF, then select All tools > Protect a PDF > Set security properties.\textbackslash n2. In the Document Properties window, select the Security tab and then select Change settings.\textbackslash n3. Your options vary [...] Select OK again to confirm the action.\\
\bottomrule
\end{tabular}
\caption{Examples of proposed method under performing DPR + GPT3.5.}
\label{table:examples2}
\end{table*}

\end{document}